\documentclass[runningheads]{llncs}
\usepackage{graphicx}
\usepackage{amsmath,amssymb} 

\usepackage{color}
\usepackage{times}
\usepackage{epsfig}
\usepackage{enumitem}

\usepackage{soul}
\newcommand{\defas}{\buildrel\triangle\over =}

\newcommand{\nn}\nonumber

\newcommand{\bc}[1]{#1}

\newcommand{\bX} { {\bf X}}

\newcommand{\bN}{ \boldsymbol{N}}

\newcommand{\cW} {\mathcal{W}}

\newcommand{\sigmoid}{Sigm}
\newcommand{\reffig}[2]{{\bf Fig.~\ref{#1}{#2}}}
\newcommand{\refeq}[1]{{\bf Eq.~\ref{#1}}}
\newcommand{\refsec}[1]{{\bf Sec.~\ref{#1}}}

\usepackage{xr}
\usepackage[pagebackref=true,breaklinks=true,letterpaper=true,colorlinks,bookmarks=false]{hyperref}
\usepackage[width=122mm,left=12mm,paperwidth=146mm,height=193mm,top=12mm,paperheight=217mm]{geometry}

\begin{document}
\pagestyle{headings}
\mainmatter
\def\ECCV16SubNumber{}  

\title{Seeing into Darkness: \\
Scotopic Visual Recognition} 
\author{Bo Chen and Pietro Perona}
\institute{California Institute of Technology, Pasadena, CA, 91125\\
\email{\{bchen3, perona\}@caltech.edu}
}

\maketitle

\begin{abstract}
Images are formed by counting how many photons traveling from a given set of directions hit an image sensor during a given time interval. When photons are few and far in between, the concept of `image' breaks down and it is best to consider directly the flow of photons. Computer vision in this regime, which we call `scotopic', is radically different from the classical image-based paradigm in that visual computations (classification, control, search) have to take place while the stream of photons is captured and decisions may be taken as soon as enough information is available. The scotopic regime is important for biomedical imaging, security, astronomy and many other fields. Here we develop a framework that allows a machine to classify objects with as few photons as possible, while maintaining the error rate below an acceptable threshold. A dynamic and asymptotically optimal speed-accuracy tradeoff is a key feature of this framework. We propose and study an algorithm to optimize the tradeoff of a convolutional network directly from lowlight images and evaluate on simulated images from standard datasets. Surprisingly, scotopic systems can achieve comparable classification performance as traditional vision systems while using less than $0.1\%$ of the photons in a conventional image. In addition, we demonstrate that our algorithms work even when the illuminance of the environment is unknown and varying. Last, we outline a spiking neural network coupled with photon-counting sensors as a power-efficient hardware realization of scotopic algorithms. 
\end{abstract}

\section{Introduction}

\label{sec-intro}

Vision systems are optimized for speed and accuracy.
Speed depends on the time it takes to capture an image (exposure time) and the time it takes to compute the answer.
Computer vision researchers typically assume that  there is plenty of light and a large number of photons may be collected very quickly\footnote{In images with 8  bits per pixel of signal (i.e. SNR=256) pixels collect $10^4-10^5$ photons~\cite{morris2015imaging}. In full sunlight the exposure time is about 1/1000 s which is negligible compared to typical computation times.}, thus speed is limited by computation. 
This is called {\em photopic vision} where the image, while difficult to interpret, is (almost) noiseless; researchers ignore exposure time and focus on the trade-off between accuracy and computation time (e.g. Fig 10 of \cite{dollarABP14}).  


\begin{figure}
\centering
\includegraphics[width=0.8\linewidth]{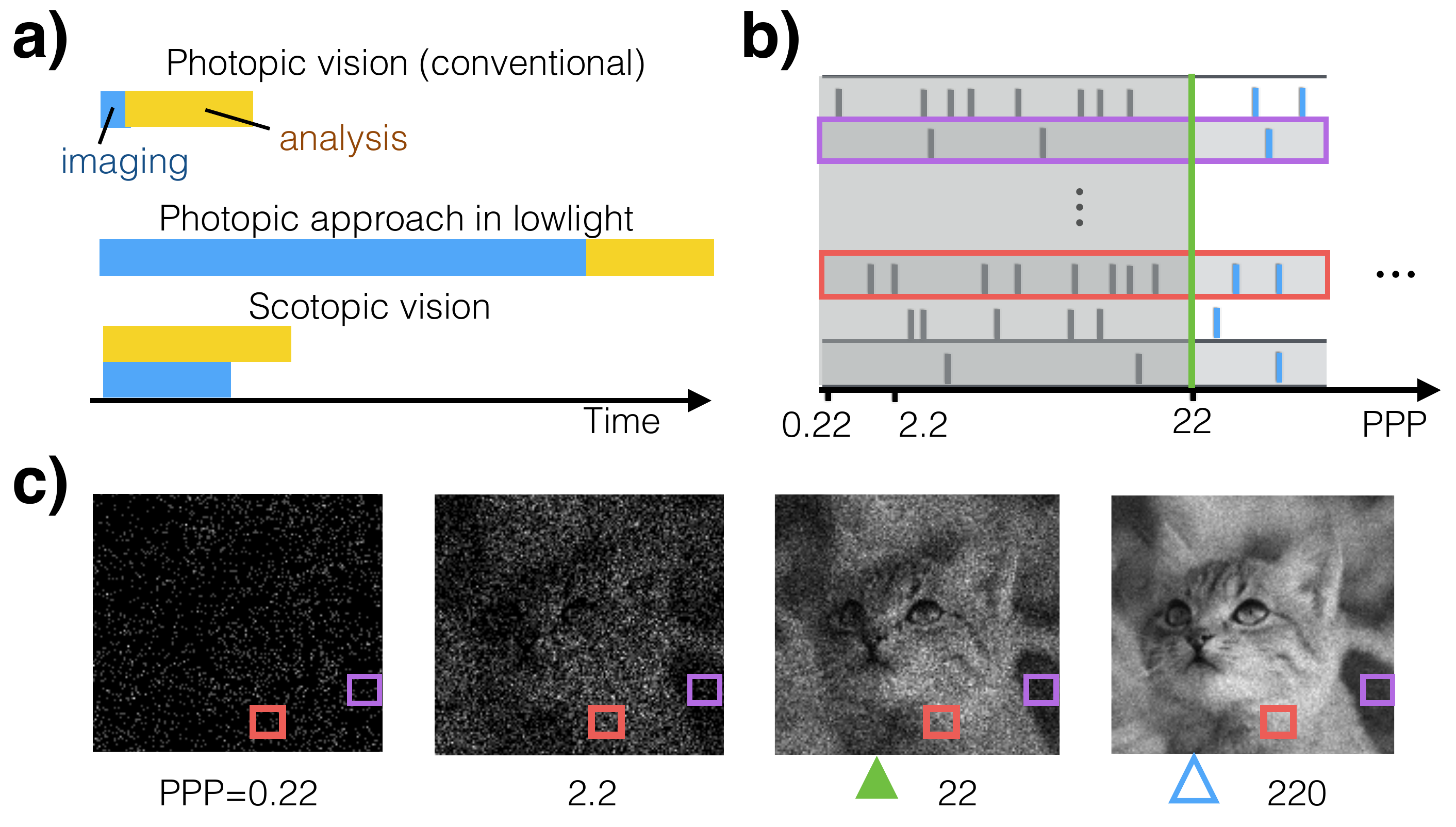}
\vspace{-0.2in}
\caption{{\bf Scotopic visual classification.} 
{\scriptsize {\bf a)} Computation time breakdown of photopic vs scotopic approaches. In conventional photopic approaches, image formation time ($\approx 30$ms) is dwarfed by the time spent in analysis (preprocessing, running classifiers, etc). The same approach in an environment that's $100\times$ darker (e.g. twilight) is confronted with a substantial slowdown due to prolonged imaging time. The proposed scotopic approach reduces runtime by 1) analyzing input as photons stream in and 2) terminating photon collection as soon as sufficient information has been collected for the particular input. {\bf b)} A sample photon stream. Each row coresponds to one pixel and each vertical bar is a photon arrival event (color-coded rows correspond to marked pixels in c)). The `amount of light' that has been collected is quantified by the average photons per pixel (PPP), which is proportional to the exposure time $t$ assuming constant illuminance. {\bf c)} Cumulative photon counts at selective PPPs visualized as images. Blue hollow arrow indicates a typical median PPP required for our scotopic classifier (WaldNet,~\refsec{sec:waldnet}) to achieve a comparable error rate as the model trained and tested using images under normal lighting conditions with about $2^7 \approx 10^4$ PPP (see~\refsec{sec:datasets} for protocol). Green solid arrow (and bar in b)) indicates the median PPP to stay within $1\%$ performance degradation.}
}
\label{fig-lowlightimages}
\vspace{-.2in}
\end{figure}

Consider now the opposite situation, which we call   {\em scotopic vision}\footnote{The term `scotopic / photopic vision' literally means `vision in the dark / with plenty of light'. It is usually associated to the physiological state where only rods, not cones, are active in the retina. We use `scotopic vision' to denote the general situation where a visual system is starved for photons, regardless the technology used to capture the image.}, where photons are few and precious, and exposure time is long compared to computation time.  The design tradeoff is between accuracy and exposure time~\cite{ferree1929intensity}, and computation time becomes a small additive constant. 

Why worry about scotopic vision? We ask the opposite question: {\em ``Why waste time collecting unnecessary photons?''} There are multiple situations where this question is compelling. (1) One may be trying to sense/control dynamics that are faster than the exposure time that guarantees good quality pictures, e.g. automobiles and quadcopters~\cite{
dickmanns2007dynamic}. (2) In competitive scenarios, such as sports, a fraction of a second may make all the difference between defeat and victory~\cite{thorpe1996speed}. (3) Sometimes prolonged imaging has negative consequences, e.g. because phototoxicity and bleaching alter a biological sample~\cite{stephens2003light} or because of health risks in medical imaging~\cite{hall2014cancer}. (4) In sensor design, reduced photon counts allow for imaging with smaller pixels and ultra-high resolution~\cite{sbaiz2009gigavision,fossum2011quanta}. (5) sometimes there is little light in the environment, e.g. at night, and obtaining a good quality image takes a long time relative to achievable computational speed. Thus, it is compelling to understand how many photons are needed for good-enough vision, and how one can make visual decisions as soon as a sufficient number of photons has been collected. In scotopic vision photons are collected until the evidence is sufficient to make a decision.

Our work is further motivated by the recent development of {\it photon-counting imaging sensors}: single photon avalanche diode arrays~\cite{zappa2007principles}, quanta image sensors~\cite{fossum2011quanta}, and gigavision cameras~\cite{sbaiz2009gigavision}. \bc{These sensors detect and report {\bf single photon arrival events} in quick succession, an ability that provides fine-grained control over photon acquisition that is ideal for scotopic vision applications. By contrast, conventional cameras, which are designed to return a high-quality image after a fixed exposure time, produce an insurmountable amount of noise when forced to read out images rapidly and are suboptimal at low light. Current computer vision technology has not yet taken advantage of photon-counting sensors since they are still under development. Fortunately, realistic noise models of the sensors~\cite{fossum2011quanta} are already available, making it possible (and wise) to innovate computational models that leverage and facilitate the sensor development.}

While scotopic vision has been studied in the context of the physiology and technology of image sensing~\cite{barlow1962method,
delbruck1994analog}, as well as the physiology and psychophysics of visual discrimination~\cite{gold2002} and visual search~\cite{chenNP11}, little is known regarding the computational principles for high-level visual tasks, such as categorization and detection, in scotopic settings. Prior work on photon-limited image classification~\cite{wernick1986image} deals with a single image, and does not study the trade-off between exposure time and accuracy. \bc{Instead, our work explores scotopic visual categorization on modern datasets such as MNIST and CIFAR10~\cite{krizhevsky2009learning,lecun1998gradient}, examines model performance under common sensory noise, and proposes realistic implementations of the algorithm. }
%

Our main contributions are:\\
1. A {\bf computational framework} for scotopic classification that can trade-off accuracy and response time. \\
2. A feedforward architecture yielding {\bf any-time, quasi-optimal} scotopic classification.\\
3. A {\bf learning algorithm} optimizing the speed accuracy tradeoff of lowlight classifiers.\\
4. {\bf Robustness analysis} regarding sensor noise in current photon-counting sensors. \\
\bc{5. A {\bf spiking implementation} that trades off accuracy with computation / power. \\
6. A {\bf light-level estimation} capacity that allows the implementation to function without an external clock and at situations with unknown illuminance levels. }



\section{Previous Work}
Our approach to scotopic visual classification is probabilistic. At every time instant each classification hypothesis is tested based on the available evidence. Is there sufficient evidence to make the decision? If so, then the pattern is classified. If not, the system will delay classification, acquire more photons, i.e. more information, and try again. This approach descends from Wald's Sequential Probablistic Ratio Test (SPRT)~\cite{wald1945sequential}. Wald proved optimality of SPRT under fairly stringent conditions (see~\refsec{sec-framework}). Lorden, Tartakowski and collaborators~\cite{lorden1977nearly,tartakovsky1998asymptotic} later showed that SPRT is {\em quasi-optimal} in more general conditions, such as the competition of multiple one-sided tests, which turns out to be useful in multiclass visual classification.

Feedforward convolutional neural networks (ConvNets)~\cite{fukushima1980neocognitron} have been recently shown to be trainable to classify images with great accuracy~\cite{lecun1998gradient,krizhevsky2012imagenet,jia2014caffe}. We show that ConvNets are inadequate for scotopic vision. However, they are very appropriate once opportune modifications are applied. In particular, our scotopic algorithm marries ConvNet's specialty for classifying good-quality images with SPRT's ability to trade off photon acquisition time with classification accuracy in a near-optimal fashion. 

Sequential testing has appeared in the computer vision literature~\cite{viola2001rapid,moreels2004recognition,matas2005randomized} in order to {\em shorten computation time}.  These algorithms assume that all visual information (`the image') is present at the beginning of computation, thus focus on reducing computation time in photopic vision. By contrast, {\em our work aims to reduce capture time} and is based on the assumption that computation time is negligible when compared to image capture time. \bc{In addition, these algorithms either require an computationally intensive numerical optimization~\cite{naghshvar2013active} or fail to offer optimality guarantees~\cite{zhu2014active,chen2014hierarchical}. In comparison, our proposed strategy has a closed-form and is asymptotically optimal in theory.}

\bc{Sequential reasoning has seen much recent success thanks to the use of recurrent neural networks (RNN)~\cite{hochreiter1997long,graves2009novel}. Our work is inherently recurrent as every incoming photon prompts our system to update its decision, hence we exploits recurrence for efficient computation. Yet, conventional RNNs are trained with inputs that are sampled uniformly in time, which in our case would translate to $>1k$ photon counting images per second and would be highly inefficient. Instead, we employ a continuous-time RNN~\cite{li2005approximation} approximation that can be trained using images sampled at arbitrary times, and find that a logarithmic number of ($4$) samples per sequence suffice in practice. }

Scotopic vision has been studied by physiologists and psychologists. Traditionally their main interest is understanding the physiology of phototransduction and of local circuitry in the retina as well as the sensitivity of the human visual system at low light levels~\cite{westheimer1965spatial,frumkes1972rod,atick1992does}, thus there is no attempt to understand `high level' vision in scotopic conditions. More recent work has begun addressing visual discrimination and search under time-pressure, such as phenomenological diffuse-to-threshold models~\cite{ratcliff1985theoretical} and Bayesian models of discrimination and of visual search~\cite{gold2002,drugowitsch2012cost,chenNP11}. The pictures used in these studies are the simple patterns (moving gratings and arrangements of oriented lines) that are used by visual psychophysicists. Our work is the first attempt to handle general realistic images.

Lastly, VLSI designers have produced circuits that can signal pixel `events' asynchronously~\cite{delbruck1993silicon,delbruck1994analog,liu2014event} as soon as a sufficient amount of signal is present. This is ideal for our work since conventional cameras acquire images synchronously (all pixels are shuttered and A-D converted at once) and are therefore ill-adapted to scotopic vision algorithms. The idea of event-based computing has been extended to visual computation by O'Connor et al.~\cite{o2013real} who developed an event-based deep belief network that can classify handwritten digits. The classification algorithms and the spiking implementation that we propose are distantly related to this work. Our emphasis is to study the best strategies to minimize response time, while their emphasis is on spike-based computation.

\section{A Framework for Scotopic Classification}
\label{sec-framework}

\subsection{Image Capture}

Our computational framework starts from a model of image capture. Each pixel in an image reports the brightness estimate of a cone of visual space by counting photons coming from that direction. The estimate improves over time. 
Starting from a probabilistic assumption of the imaging process and of the target classification application, we derive an approach that allows for the best tradeoff between exposure time and classification accuracy.

We make three assumptions and relax them later:
\begin{itemize}
\item The world is stationary during the imaging process. This may be justified as many photon-counting sensors sample the world at $>1kHz$~\cite{sbaiz2009gigavision,fossum2011quanta}. \bc{Later we test the model under different camera movements and show robust performance.}

\item Photon arrival times follow a homogeneous Poisson process. This assumption is only used to develop the model. We will evaluate the model in~\refsec{sec:noise} using observations from realistic noise sources. 
\item A probabilistic classifier based on photon counts is available. We discuss how to obtain such a model in~\refsec{sec:learning}.
\end{itemize}

Formally, the input $\bX_{1:t}$ is a stream of photons incident on the sensors during time $[0, t\Delta)$, where time has been discretized into bins of length $\Delta$. $X_{t,i}$ is the number of photons arrived at pixel $i$ in the $t$th time interval, i.e. $[(t-1)\Delta, t\Delta)$. \bc{The task of a scotopic visual recognition system is two fold: 1) computing the category $C\in \{0,1,\ldots,K\}$ of the underlying object, and 2) crucially, determining and minimizing the exposure time $t$ at which the observations are deemed sufficient. }

\subsubsection{Noise Sources} \label{sec:noises}
The pixels in the image are corrupted by several noise sources intrinsic to the camera~\cite{liu2008automatic}. {\bf Shot noise}: The number of photons incident on a pixel $i$ in the unit time follows a Poisson distribution whose rate $\lambda_i$ (Hz) depends on both the pixel intensity $I_i\in[0,1]$ and a {\bf dark current} $\epsilon_{dc}$:
\begin{align}
P(X_{t,i}=k) = Poisson(k|\lambda_i t) = Poisson(k | \lambda_\phi\frac{ I_i+\epsilon_{dc}}{1+\epsilon_{dc}} t) \label{eq:noise}
\end{align}
 where $\lambda_\phi$ is the illuminance (maximum photon count per pixel) per unit time~\cite{morris2015imaging,sbaiz2009gigavision,liu2008automatic,fossum2013modeling}. During readout, the photon count is additionally corrupted first by the amplifier's {\bf read noise}, which is an additive Gaussian, then by the {\bf fixed-pattern noise} which may be thought of as a multiplicative Gaussian noise~\cite{healey1994radiometric}. As photon-counting sensors are designed to have low read noise and low fixed pattern noise\cite{fossum2011quanta,zappa2007principles,fossum2013modeling}, we focus on modeling the shot noise and dark current only. We will show (\refsec{sec:noise}) that our models are robust against all four noise sources. Additionally, according to the stationary assumption there is no {\it motion-induced blur}. For simplicity we do not model {\it charge bleeding and cross-talk} in colored images, and assume that they will be mitigated by the sensor community~\cite{anzagira2015color}.



When the illuminance $\lambda_{\phi}$ of the environment is fixed, the average number of photons per pixel (PPP)\footnote{PPP is averaged across the entire scene and duration.} is linear in $t$:
\begin{align}
PPP&=\lambda^{\phi}t\Delta \label{eq:PPP}
\end{align}
 hence we will use time and PPP interchangeably. Since the information content in the image is directly related to the amount of photons, from now on we measure response time in terms of PPP instead of exposure time. \reffig{fig-lowlightimages}{} shows a series of images with increasing PPP. \bc{See~\refsec{sec:lightestimate} for cases where the illuminance is varying in time.}

\subsection{Sequential probability ratio test} \label{sec:sprt}
Our decision strategy for trading off accuracy and speed is based on SPRT, for its simplicity and attractive optimality guarantees. Assume that a probabilistic model is available to predict the class label $C$ given a  sensory input $\bX_{1:t}$ of any duration $t$ -- either provided by the application or learned from labeled data using techniques described in~\refsec{sec:learning} -- SPRT conducts a simple accumulation-to-threshold procedure to estimate the category $\hat{C}$: 

Let $S_c(\bX_{1:t})\defas \log \frac{P(C=c|\bX_{1:t})}{P(C\neq c|\bX_{1:t})}$ denote the class posterior probability ratio of the visual category $C$ for photon count input $\bX_{1:t}$, $\forall c\in\{1,\ldots,K\}$, and let $\tau$ be an appropriately chosen threshold. SPRT conducts a simple accumulation-to-threshold procedure to estimate the category $\hat{C}$:
\begin{align}
	\mbox{ Compute } &c^* = \underset{c=1,\ldots,K}{\arg\max}S_c(\bX_{1:t}) \nn \\
	\mbox{ if } S_{c^*}(\bX_{1:t}) &> \tau : \mbox{ report } \hat{C}=c^*  \nn\\
	\mbox{ otherwise }: & \mbox{ increase exposure time $t$}.  \label{eq:sprt}
\end{align}

\begin{figure}[t]
\begin{centering}
\includegraphics[width=0.8\linewidth]{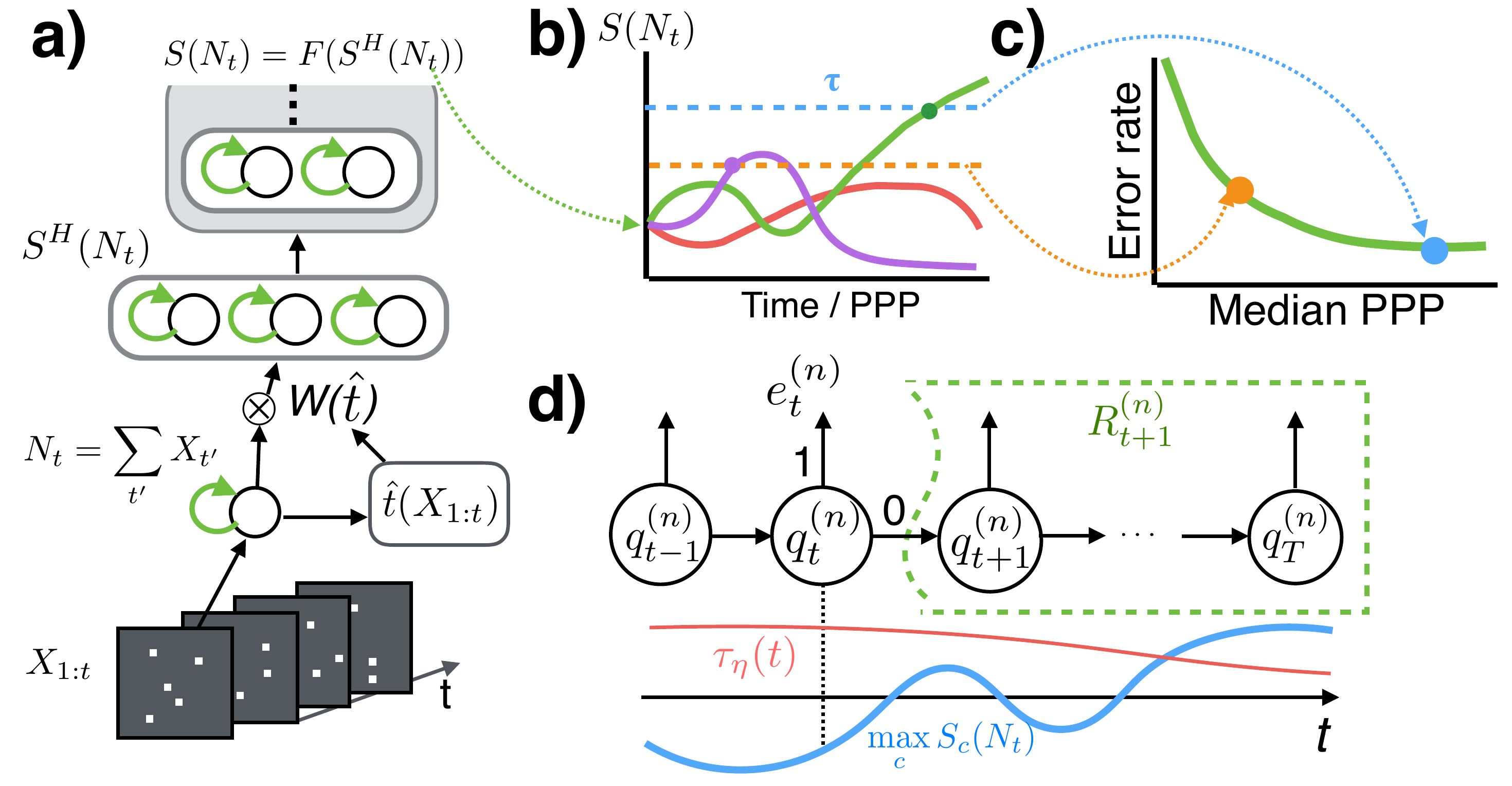}
\vspace{-0.1in}
\caption{ {\bf WaldNet for lowlight visual recognition.} {\scriptsize {\bf(a)} Computing class posterior ratios. The first layer is adapted (\refeq{eq:approxh}) to capture time-invariant features $S^H(\bN_t)$ from raw photon counts $\bX_{1:t}$ (visualization in~\reffig{fig-lowlightimages}). The key to the adaptation is to adjust the convolutional filters ${\boldsymbol{W}}$ based on exposure time $t$, assuming constant illuminance. If the illuminance is irregular and/or unpredictable, an equivalent exposure time $\hat{t}$ is estimated directly from the photon counts (\refsec{sec:lightestimate}). The first layer features feed into the remainder of the ConvNet $F$ to compute class posterior $S_c(\bN_t)=\frac{P(C=c|\bX_t)}{P(C\neq c|\bX_t)}$. $S_c(\bN_t)$ may be computed efficiently using a spiking recurrent neural network implementation (\refsec{sec:spiking}) that leverages sparsity in changes of the network's internal states. {\bf(b)} Deciding when to stop collecting photons. The class posterior ratios race to a common threshold to determine the category to report. WaldNet stops photon collection as soon as one class crosses the threshold $\tau$ (\refeq{eq:sprt}). The example shows $S_c(\bN_t)$ for three classes where the true class is green. Using a higher threshold (blue) yields a later but more accurate solution whereas a lower (orange) threshold is faster but risks misclassification. {\bf (c)} A speed versus accuracy tradeoff curve (illustration only) produced by repeating (a-b) for multiple images and sweeping the threshold $\tau$.{\bf (d)}
Learning time-varying threshold $\tau_\eta(t)$ (when class posterior learning (\refeq{eq:step1}) is imperfect) to optimize Bayes risk with time cost $\eta$ (\refeq{eq:bayesrisk}). The centipede network describes the recurrence relationship between risk $R_t^{(n)}$ starting from time $t$ of example $n$ and the risk $R_{t+1}^{(n)}$ starting from time $t+1$ (\refeq{eq:R_recursive}). $q^{(n)}_t$ is a gate (based on whether $\max_cS_c(\bN_t)$ crosses threshold) that decides whether WaldNet stops at $t$ with misclassification risk $e^{(n)}_t$ or continue collecting photons with risk $R_{t+1}^{(n)}$. }
}
\label{fig-model}
\end{centering}
\vspace{-.2in}
\end{figure}

When a decision is made, the declared class $\hat{C}$ has a probability that is at least $\exp(\tau)$ times bigger thanthe probability of all the other classes combined, therefore the error rate of SPRT is at most $1-\sigmoid(\tau)$, where $\sigmoid$ is the sigmoid function: $\sigmoid(x)\defas \frac{1}{1+\exp(x)}$. 

For simple binary classification problems, 
SPRT is optimal in trading off speed versus accuracy in that no other algorithm can respond faster while achieving the same accuracy~\cite{wald1945sequential}. In the more realistic case where the categories are rich in intra-class variations, SPRT is shown to be asymptotically optimal, i.e. it gives optimal error rates as the exposure time becomes large~\cite{lorden1977nearly}. Empirical studies suggest that even for short exposure times SPRT is near-optimal~\cite{chen2014towards}. 

In essence, SPRT decides when to respond dynamically, based on the stream of observations accumulated so far. Therefore, the response time is different for each example. This regime is called ``{\bf free-response}'' (FR), in contrast to the ``{\bf interrogation}'' (INT) regime, typical of photopic vision, where a fixed-length observation is collected for each trial~\cite{bogacz2006physics}. 
The observation length may be chosen according to a training set and fixed a priori. In both regimes, the length of observation should  take into account the cost of errors, the cost of time, and the difficulty of the classification task. 

\bc{Despite the striking similarity between the two regimes, SPRT (the FR regime) is provably optimal in the asymptotical tradeoff between response time and error rate, while such proofs do not exist for the INT regime. We will empirically evaluate both regimes in~\refsec{sec:fr}.}

\subsection{Computing class probabilities over time}
The challenge of applying SPRT is to compute $S_c(\bX_{1:t})$ for class $c$ and the input stream $\bX_{1:t}$ of variable exposure time $t$, or in a more information-relevant unit, variable PPP levels. Thanks to the Poisson noise model (\refeq{eq:noise}), the sufficient statistics for observation $\bX_{1:t}$ is the cumulative count $\bN_t=\sum_{t'=1}^t\bX_{t'}$ (visualized in ~\reffig{fig-lowlightimages}{}), and the observation duration length $t$, therefore we may rewrite $S_c(\bX_{1:t})$ as $S_c(\bN_t, t)$. We further shorthand the notation to $S_c(\bN_t)$ since the exposure time is evident from the subscript. Since counts at different PPPs (and, equivalently, exposure times) have different statistics, it would appear that a specialized system is required for each PPP. This leads to the naive \emph{ensemble} approach. Instead, we also propose a network called \emph{WaldNet} that can process images at all PPPs and has the size of only a single specialized system. We describe the two approaches below. 

\subsubsection{Model-free approach: network ensembles}
\label{sec-naive}
The simple idea is to build a separate `specialist' model $S(\bN_{t})$ for each exposure time $t$ (or light level PPP), either based on domain knowledge or learned from a training set. For best results one needs to select a list of representative light levels to train the specialists, and route input counts $\bN_t$ to the specialist closest in light level. We refer to this as the `ensemble' predictor. 

One potential drawback of this ensemble approach is that training and storing multiple specialists is \emph{wasteful}. At different light levels, while the cumulative counts change drastically, the underlying statistical structure of the task stays the same. An approach that takes advantage of this relationship may lead to more parsimonious algorithms.

\subsubsection{Model-based approach: WaldNet}\label{sec:waldnet}
Unlike the ensemble approach, we show that one can exploit the structure of the input and build one system for images at all PPPs. The variation in the input $\bN_t$ has two independent sources: one is the stochasticity in the photon arrival times, and the other the intra- and inter- class variation of the real intensity values of the object. SPRT excels at reasoning about the first noise source while deep networks are ideal for capturing the second. 
Therefore we propose WaldNet, a deep network for speed-accuracy tradeoff (\reffig{fig-model}{b-c}) that combines ConvNets with SPRT. \bc{WaldNet automatically adjusts the parameters of a ConvNet according to the exposure time $t$. Thus a WaldNet may be viewed as an ensemble of infinitely many specialists that occupies the size of only one specialist.} 


The key ingredient to SPRT is the log posterior ratios $S(\bN_t)$ over exposure time $t$. Standard techniques such as ConvNets can not be applied directly as their input $\bN_T$ is static, namely the cumulative photon counts up to an identical exposure time $T$ (e.g. $T\Delta \approx33ms$ in normal lighting conditions). However we propose a simple adjustment that transfers the uncertainty in the photon counts to uncertainty in the task-relevant features of a ConvNet. 

A standard ConvNet contains multiple layers of computations that may be viewed as a nesting of two transformations: (1) the first hidden layer $S^H(\bN_T)=\textbf{ W}\bN_T+\textbf{ b}^H$ that maps the input to a feature vector\footnote{\bc{Without loss of generality and for notational simplicity, we assume that the first layer is fully-connected as oppose to convolutional. \refsec{S-app:approxh} discusses how to extend the results here to convolutional layers. We also define the first layer feature as the activity prior to non-linearity.}}, and (2) the remaining layers $S(\bN_T) = F(S^H(\bN_T))$ that map the features $S^H(\bN_T)$ to the log class posterior probabilities $S(\bN_T)$. $\textbf{ W}\in \mathbb{R}^{D\times n_H}$ is a weight vector and $\textbf{ b}^H \in \mathbb{R}^{n_H}$ is a bias vector.
%



Given only partial observations $\bN_t$, computing features of the first layer requires marginalizing out unobserved photon counts $\Delta \bN\defas \sum_{t'=t+1}^T\bX_{t'}$. The marginalization requires putting a prior on the photon emission rate per image pixel $i$, which we assume to be a Gamma distribution: $Gam(\mu_it_0, t_0)$, where $\mu_i$ represents the prior mean rate for pixel $i$ and $t_0$ (shared across pixels) represents the strength of the prior\footnote{We use a Gamma prior because it is the conjugate prior of the Poisson likelihood.}. Then the first layer of hidden features may be approximated by:
\begin{align}
	S^H(\bN_t) &= \sum_{\Delta \bN} (\boldsymbol{W}(\bN_t + \Delta \bN)+\boldsymbol{b}^H) P(\Delta \bN|\bN_t)\approx \alpha(t)\boldsymbol{W}\bN_t + \boldsymbol{\beta}(t) \label{eq:approxh} 
\end{align}

where the scaling factor $\alpha(t)\defas \frac{T+t_0}{t+t_0}$ is a scalar and the biases $\boldsymbol{\beta}(t)$ is a length $n_H$ vector. For the $j$-th hidden feature, $\beta_j(t) \defas \frac{t_0(T-t)}{t+t_0} \sum_i W_{ij}  \mu_i + b_j $. Detailed derivations are in~\refsec{S-app:approxh}.

Essentially, the adaptation procedure in~\refeq{eq:approxh} accounts for the stochasticity in photon arrival time by using time-dependent weights and biases, rendering an exposure-time invariant feature representation $S^H(\bN_t)$. The computations downstream, $F$, may then treat $S^H(\bN_t)$ as if it were obtained from the entire duration. Therefore the same computations suffice to model the intra- and inter-class variations: $S(\bN_t) = F(S^H(\bN_t))$.

 The network is trained discriminately (\refsec{sec:learning}) with the first layer replaced by~\refeq{eq:approxh}. 
The network has nearly the same number of parameters as a conventional ConvNet, but has the capacity to process inputs at different exposure times. The adaptation is critical for performance, as will be seen by comparison with simple rate-based methods in~\refsec{sec:exp}. See~\refsec{sec:waldnet-imp} for implementation details. 


\subsection{Learning} \label{sec:learning}


Our goal is to train WaldNet to optimally trade off the expected exposure time (or PPP) and error rate in the FR regime. Optimality is defined by the Bayes risk $R$~\cite{wald1945sequential}:
\begin{align}
	R &\defas \eta \mathbb{E}[\mbox{PPP}] + \mathbb{E}[C\neq \hat{C}] \label{eq:bayesrisk}
\end{align}
where $\mathbb{E}$[PPP] is the expected photon count required for classification, $\mathbb{E}[C\neq \hat{C}]$ is the error rate, and $\eta$ describes the user's cost of photons per pixel (PPP) versus error rate. WaldNet asymptotically optimizes the Bayes risk provided that it can faithfully capture the class log posterior ratio $S_c(\bN_t)$, and selects the correct threshold $\tau$ (\refeq{eq:sprt}) based on the tradeoff parameter $\eta$. Sweeping $\eta$ traverses the optimal time versus error tradeoff (\reffig{fig-model}{c}). 

Since picking the optimal threshold according to $\eta$ is independent from training a ConvNet to approximately compute the log posterior ratio $S_c(\bN_t)$, the same ConvNet is shared across multiple $\eta$'s. This suggests the following two-step learning algorithm. 

\vspace{-.2in}
\subsubsection{Step one: posterior learning }
Given a dataset $\{\bN^{(n)}_t, C^{(n)}\}_{n, t}$ where $n$ indexes training examples and $t$ indexes exposure time, we train the adapted ConvNet to minimize:
\begin{align}
 - \sum_{n,t} \log P(\hat{C}=C^{(n)} | \bN^{(n)}_t, \cW)  + reg(\cW) \label{eq:step1}
\end{align}
where $\cW$ collectively denote all the parameters in the adapted ConvNet, and $reg(\cW)$ denotes $L2$ weight-decay on the filters. When a lowlight dataset is not available we simulate the dataset from normal images according to the noise model in~\refeq{eq:noise}, where the exposure times are sampled uniformly on a logarithmic scale (see \refsec{sec:exp}).

\vspace{-.2in}
\subsubsection{Step two: threshold tuning}\label{sec:fine-tuning}
If the ConvNet in step one captures the log posterior ratio $S_c(\bN_t)$, we can simply optimize a scalar threshold $\tau_\eta$ for each tradeoff parameter $\eta$. In practice, we may opt for a time-varying threshold $\tau_\eta(t)$ as step one may not be perfect\footnote{For instance, consider an adapted ConvNet that perfectly captures the class posterior. Ignoring the regularizer (right term of~\refeq{eq:step1}), we can scale up the weights and biases of the last layer (softmax) by an arbitrary amount without affecting the error rate, which scales the negative log likelihood (left term in \refeq{eq:step1}) by a similar amount, leading to a better objective value. The magnitude of the weights are thus determined by the regularizer and may be off by a scaling factor. We therefore need to properly rescale the class posterior at every exposure time before comparing to a constant threshold, which is equivalent to using a time-varying threshold $\tau_\eta(t)$ on the raw predictions.}.  

$\tau_\eta(t)$ affects our Bayes risk objective in the following way (\reffig{fig-model}{d}). Consider a high-quality image $\bN^{(n)}_T$, let $\{\bN^{(n)}_t\}_{t=1}^T$ be a sequence of lowlight images increasing in PPP generated from $\bN^{(n)}_T$. Denote $q^{(n)}_t \defas \mathbb{I}[\max_{c} S_c(\bN_t)>\tau_\eta(t)]$ the event that the log posterior ratio crosses decision threshold at time $t$, and $e^{(n)}_t$ the event that the class prediction at $t$ is wrong. Let $R^{(n)}_t$ denote the Bayes risk of the sequence incurred from time $t$ onwards. $R^{(n)}_t$ may be computed recursively (derived in~\refsec{S-app:fine-tuning}):
\begin{align}
R^{(n)}_t &= \eta\Delta + q^{(n)}_te^{(n)}_t + (1-q^{(n)}_t) R^{(n)}_{t+1} \label{eq:R_recursive}
\end{align}
where the first term is the cost of collecting photons at time $t$, the second term is the expected cost of committing to a decision that is wrong, and the last term is the expected cost of deferring the decision till more photons are collected. 

The Bayes risk is obtained from averaging multiple photon count sequences, i.e. $R = \mathbb{E}_{(n)}[R^{(n)}_0]$. $q^{(n)}_t$ is non-differentiable with respect to the threshold $\tau_\eta(t)$, leading to difficulties in optimizing $R$. Instead, we approximate $q^{(n)}_t$ with a Sigmoid function: 
\begin{align}
q^{(n)}_t(\tau_\eta(t)) &\approx  \sigmoid\left(\frac{1}{\sigma} (\max_c S_c(\bN_t)- \tau_\eta(t)) \right) \label{eq:smooth}
\end{align}
where $\sigmoid(x)\defas 1/(1+\exp(-x))$, and anneal the temperature $\sigma$ of the Sigmoid over the course of training~\cite{mobahi2015link} (see~\refsec{sec:exp}). 

\subsection{Automatic light-level estimation}\label{sec:lightestimate}
\bc{Both scotopic algorithms (ensemble and WaldNet) assume knowledge of the light-level PPP in order to choose the right model parameters. This knowledge is easy to acquire when the illuminance is constant over time, in which case PPP is linear in the exposure time $t$ (\refeq{eq:PPP}), which may be measured by an internal clock. 

However, in situations where the illuminance is dynamic and unknown, the linear relationship between PPP and exposure time is lost. In this case we propose to estimate PPP directly from the photon stream itself, as follows. Given a cumulative photon count image $\bN$ ($t$, the time it takes to accumulate the photons, is no longer relevant as the illuminance is unknown), we examine local neighbors that receive high photon counts, and pool the photon counts as a proxy for PPP. In detail, we (1) convolve $\bN$ using an $s\times s$ box filter, (2) compute the median of the top $k$ responses, and (3) fit a second order polynomial to regress the median response towards the true PPP. Here $s$ and $k$ are parameters, which are learned from a training set consisting of $(\bN,$PPP$)$ pairs. Despite its simplicity, this estimation procedure works well in practice, as we will see in ~\refsec{sec:exp}.}

\subsection{Spiking implementation}\label{sec:spiking}
\bc{One major challenge of scotopic systems is to compute log posterior ratio computations as quickly as photons stream in. Photon-counting sensors~\cite{sbaiz2009gigavision,fossum2011quanta} sample the world at $1k-10kHz$, while the fastest reported throughput of ConvNet~\cite{ovtcharov2015accelerating} is $2kHz$ for $32\times32$ color images, and $800Hz$ for $100\times100$ color images. Fortunately, the reported throughputs are based on independent images, whereas the scotopic systems operate on photon streams, where temporal coherence may be leveraged for accelerated processing. Since the photon arrival events within any time bin is sparse, changes to the input and the internal states of a scotopic system are small. An efficient implementation thus could model the changes and propagate only those that are above a certain magnitude. 

One such implementation relies on spiking recurrent hidden units. A spiking recurrent hidden unit is characterized by two aspects: 1) computation at time $t$ reuses the unit's state at time $t-1$ and 2) only changes above a certain level will be propagated to layers above. 

Specifically, the first hidden layer of WaldNet may be exactly implemented using the following recurrent network where the features $S^H(\bN_t)$ (\refeq{eq:approxh}) are represented by membrane voltages $\boldsymbol{V}(t) \in \mathbb{R}^{n_H}$. The dynamics of $\boldsymbol{V}(t)$ is:

\begin{align}
	\boldsymbol{V}(t) = r(t) \boldsymbol{V}(t-1) + \alpha(t)\boldsymbol{W} \bX_t + \boldsymbol{l}(t) \label{eq:rnn}
\end{align}
where $r(t) \defas \frac{\alpha(t)}{\alpha(t-1)}$ is a damping factor, $\boldsymbol{l}(t) \defas \boldsymbol{\beta}(t)-r(t)\boldsymbol{\beta}(t-1)$ is a leakage term (derivations in~\refsec{S-app:snn}). The photon counts $\bX_t$ in $[(t-1)\Delta, t\Delta)$ is sparse, thus the computation $\boldsymbol{W}\bX_t$ is more efficient than computing $S^H(\bN_t)$ from scratch. 

To propagate only large changes to the layers above, we use a similar thresholding mechanism as (\refeq{eq:sprt}). For each hidden unit $j$, we associate a `positive' and a 'negative' neuron that communicate with the layer above. For each time bin $t$:
\begin{align}
	\mbox{ if } V_j(t) &> \tau_{dis} : \mbox{ send spike from positive neuron}, V_j(t)=V_j(t)-\tau_{dis}\nn\\
	\mbox{ if } V_j(t) &< -\tau_{dis} : \mbox{ send spike from negative neuron}, V_j(t)=V_j(t)+\tau_{dis} 
	\label{eq:snn}
\end{align}
where $\tau_{dis}>0$ is a discretization threshold. By taking the difference between the spike counts from the `positive' and the `negative'  neuron, the layers above can reconstruct a discretized version of $V_j(t)$. Hidden units from higher layers in WaldNet may be approximated using spiking recurrent units (\refeq{eq:rnn}) in a similar fashion.

The discretization threshold affects not only the number of communication spikes, but also the quality of the discretization, and in turn the classification accuracy. For spike-based hardwares~\cite{merolla2014million}, the number of spikes is an indirect measure of the energy consumption (Fig. 4(B) of~\cite{merolla2014million}). For non-spiking hardwares, the number of spikes also translate to the number of floating point multiplications required for the layers above. Therefore, the $\tau_{dis}$ controls the tradeoff between accuracy and power / computational cost. We will empirically evaluate this tradeoff in~\refsec{sec:exp}. }

%


\section{Experiments}
\label{sec:exp}
\vspace{-.1in}
\subsubsection{Exposure time versus signal}
Our experiments use PPP interchangeably with exposure time for performance measurement, since PPP directly relates to the number of bits of signal in each pixel. In practice an application may be more concerned with exposure time. Thus it is helpful to relate exposure time with the bits of signal. {\bf Table~\ref{table-bits}} describes this relationship for different illuminance levels (see~\refsec{S-app:time-bits} for derivations). 

\begin{table}
\centering
\begin{tabular}{| l | c|c|c|c|c|c|c|}
\hline 
 & Illuminance & \multicolumn{6}{|c|}{exposure time $t$ (s)} \\
Scene & E$_v$ (LUX) & 1/500 & 1/128 & 1/8  & 1  & 8 & 60 \\ 
\hline
\hline
Moonless & $10^{-3}$ & & & & & 1.5  & 3 \\
Full moon & $1$ &0.5 &1.5 & 3.5  & 5  & 6.5 & 8 \\
Office& $ 250 $ & 4.5  & 5.5 & 7.5 & 9 & 10.5 & 12  \\
Overcast & $10^3$ & 5.5  & 6.5  & 8.5 & 10 & 11.5 & 13  \\
Bright sun & $10^5$ & 9 & 10 & 12 & 13.5  &  15 & 16.5\\
\hline
\end{tabular}
\caption{(Approximate) number of bits of signal per pixel under different illuminance levels. See~\refsec{S-app:time-bits} for full derivation. For instance, in an office scene it takes $1/8$ seconds to obtains a $7.5$-bit image. Under full moon, the same high-quality image and the same sensor needs $>8$ seconds to capture. }
\vspace{-.3in}
\label{table-bits}
\end{table}

\subsection{Baseline Models} \label{sec:arch}

We compare WaldNet against the following baselines, under both the INT regime and the FR regime:

{\bf Ensemble}. We construct an ensemble of $4$ specialists with PPPs from $\{.22,2.2,22,220\}$ respectively. 
The performance of the specialists at their respective PPPs gives a lower bound on the optimal performance by ConvNets of the same architecture. 

 {\bf Photopic classifier}. An intuitive idea is to take a network trained in normal lighting conditions, and apply it to properly rescaled lowlight images. We choose the specialist with PPP$=220$ as the photopic classifier as it achieves the same accuracy as a network trained with $8$-bit images.


{\bf Rate classifier}. A ConvNet on the time-normalized image (rate) without weight adaptation. The first hidden layer is computed as $S^H_j(\bN_t) \approx \boldsymbol{W}\bN_t/t + \boldsymbol{b}^H $. Note the similarity with the WaldNet approximation used in~\refeq{eq:approxh}.

\bc{For all models above we assume that an internal clock measures the exposure time $t$, and the illuminance $\lambda^\phi$ is known and constant over time. We remove this assumption for the model below for unknown and varying illuminance:

{\bf WaldNet with estimated light-levels}. A WaldNet that is trained on constant illuminance $\lambda^\phi$, but tested in environments with unknown and dynamic illuminance. In this case the linear relationship between exposure time $t$ and PPP (\refsec{eq:PPP}) is lost. Instead, the light-level is first estimated according to~\refsec{sec:lightestimate} directly from the photon count image $\bN$. The estimate $\hat{PPP}$ is then converted to an `equivalent' exposure time $\hat{t}$ using $\hat{t} =\frac{ \hat{PPP} }{\lambda^\phi\Delta}$ (by inverting~\refeq{eq:PPP}), which is used to adapt the first hidden layer of WaldNet in~\refsec{eq:approxh}, i.e. $S^H(\bN) \approx \alpha(\hat{t})\boldsymbol{W}\bN + \boldsymbol{\beta}(\hat{t})$.  }

%
%
%
%

\subsection{Datasets}\label{sec:datasets}
We consider two standard datasets: MNIST~\cite{lecun1998gradient} and CIFAR10~\cite{krizhevsky2009learning}. We simulate lowlight image sequences using~\refeq{eq:noise}.

MNIST contains gray-scaled $28\times28$ images of $10$ hand-written digits. It has $60k$ training and $10k$ test images. We treat the pixel values as the ground truth intensity\footnote{The brightest image we synthesize has about $2^8$ photons, which corresponds to a pixel-wise maximum signal-to-noise ratio of $16$ ($4$-bit accuracy), whereas the original MNIST images has ($7$ to $8$-bit accuracy) that corresponds to $2^{14}$ to $2^{16}$ photons. }. Dark current $\epsilon_{dc}=3\%$. We use the default LeNet architecture from the MatConvNet package~\cite{vedaldi15matconvnet} with batch normalization~\cite{ioffe15batch} after each convolution layer. The architecture is $784$-$20$-$50$-$500$-$10$\footnote{The first and last number represent the input and output dimension, each number in between represents the number of feature maps used for that layer. The number of units is the product of the number of features maps with the size of the input.} with $5\times 5$ receptive fields and $2\times 2$ pooling. 

CIFAR10 contains $32\times32$ color images of $10$ visual categories. It has $50k$ training and $10k$ test images. We use the same sythensis procedure above to each color channel\footnote{For simplicity we do not model the Bayer filter mosaic.}. We again use the default $1024$-$32$-$32$-$64$-$10$ LeNet architecture~\cite{krizhevsky2012imagenet} with batch normalization. We use the same setting prescribed in~\cite{krizhevsky2012imagenet} to achieve $18\%$ test error on normal lighting conditions. \cite{krizhevsky2012imagenet} uses local contrast normalization and ZCA whitening as preprocessing steps. We estimate the local contrast and ZCA from normal lighting images and transforming them according to the lowlight model to preprocess scotopic images. 

\subsubsection{Training}
\bc{We train all models for MNIST and CIFAR10 using stochastic gradient descent with mini-batches of size $100$. For MNIST, we use $5k$ training examples for validation and train on the remaining $55k$ examples for $80$ iterations. We found that empirically a learning rate of $0.004$ works best for WaldNet, and $0.001$ works best for the other architectures. As CIFAR10 is relatively data-limited, we do not use a validate set and instead train all models for $75$ epochs, where the learning rate is $0.05$ for $30$ iterations, $0.005$ for other $25$ then $0.0005$ for the rest. Again, quadrupling the learning rate empirically improves WaldNet's performance but not the other architectures.}

\begin{figure}[t]
\begin{centering}
\hspace{-0.2in} \includegraphics[width=0.8\linewidth]{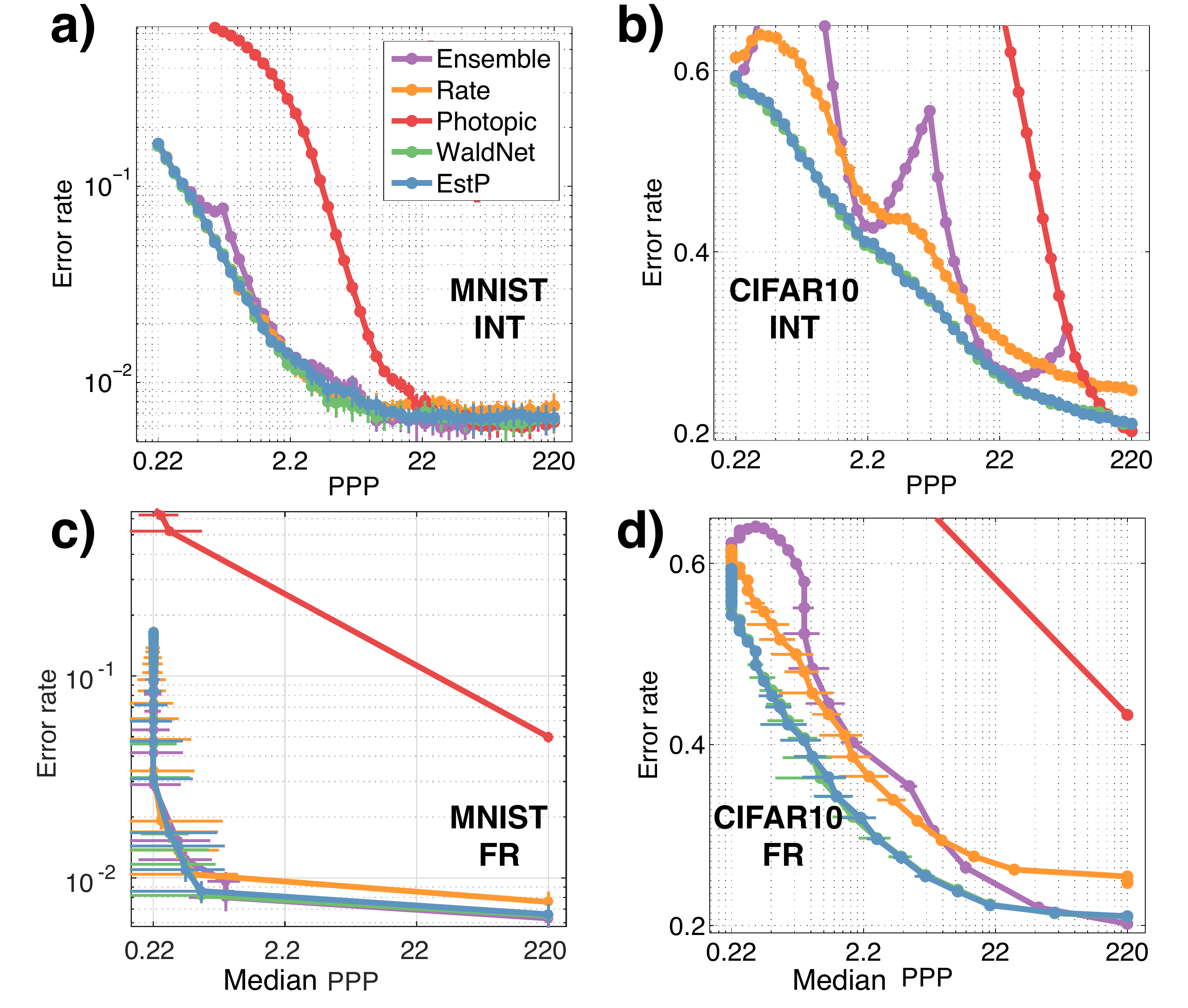}
\vspace{-0.1in}
\caption{ {\bf Performance comparison}. ({\bf a,b)}) Error rate plotted against the interrogation PPP for ({\bf a}) MNIST and ({\bf b}) CIFAR10. Each dot is computed from classifying $10k$ test examples with a fixed PPP. ({\bf c,d)}) Error rate plotted against {\it median} PPP for ({\bf c}) MNIST and ({\bf d}) CIFAR10. $1$ bootstrap {\it ste} is shown for both the median PPP and error rate, the latter is too small to be visible.}
\label{fig-SATwaldnet}
\end{centering}
\vspace{-.1in}
\end{figure}

\begin{figure}[t]
\begin{centering}
\includegraphics[width=0.8\linewidth]{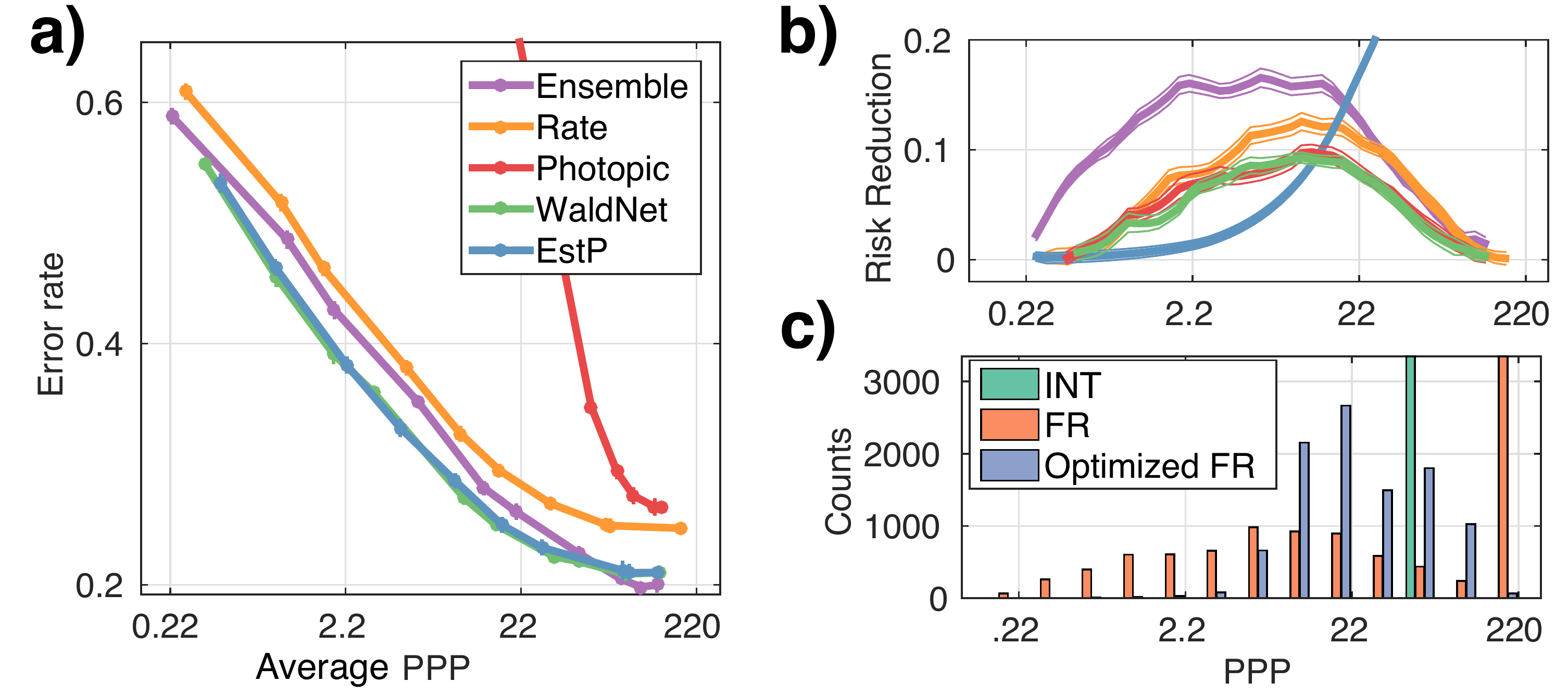}
\vspace{-0.2in}
\caption{ {\bf The effect of threshold learning (\refsec{sec:fine-tuning})}. ({\bf a}) Error rate against the {\it average} PPP for CIFAR10 using a network with optimized time-varying threshold $\tau_\eta(t)$. $1$ bootstrapped {\it ste} is shown but not visible. ({\bf b}) Each curve shows the Bayes risk reducation after optimization (\refsec{sec:learning}, step two) per {\it average} PPP. ({\bf c}) Response time (PPP) histograms under INT, FR (before optimization), and FR (after optimization) of a WaldNet that achieves $22\%$ error on CIFAR10. }
\vspace{-.2in}
\label{fig-SATwaldnet-finetuned-mean}
\end{centering}
\end{figure}

\begin{figure}[t]
\begin{centering}
 \includegraphics[width=1\linewidth]{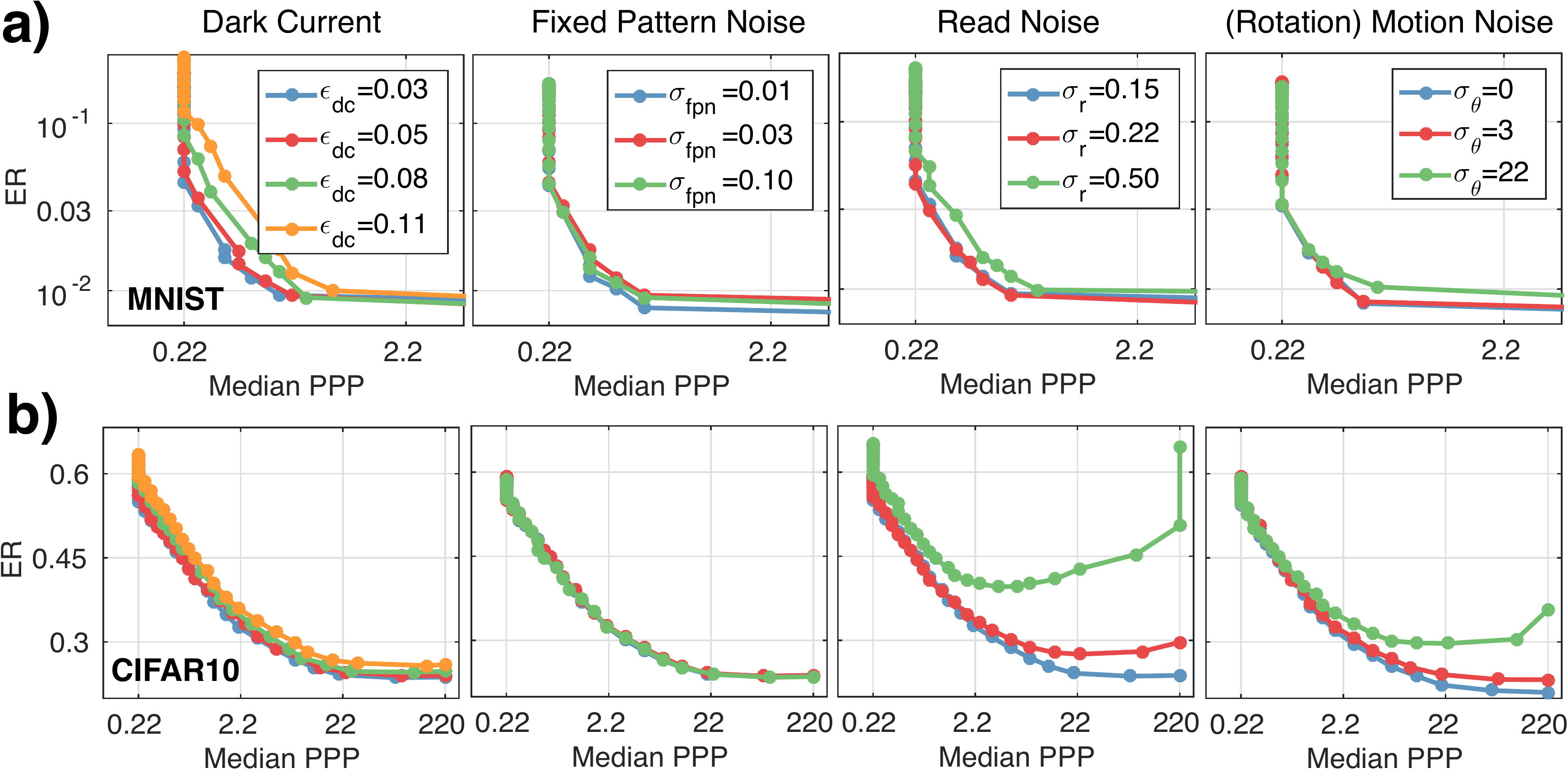}
 \vspace{-0.2in}
\caption{ {\bf The effect of sensor noise on WaldNet}. The rows correspond to datasets MNIST and CIFAR10, and the columns correspond to parameters of noise sources, which are the dark current $\epsilon_{dc}$, the standard deviation of multiplicative fixed pattern noise $\sigma_{fpn}$, the std of additive read noise $\sigma_{r}$, and the std of the rotational jitter $\sigma_\theta$ in degrees. Only one noise is varied in each panel while the rest are fixed at their respective baseline: $\epsilon_{dc}=3\%$,$\sigma_{r}=0.15, \sigma_{fpn}=3\%$ and $\sigma_\theta=0$. }
\label{fig-SATwaldnet-noise}
\end{centering}
\vspace{-.2in}
\end{figure}
\subsubsection{Implementation of WaldNet}\label{sec:waldnet-imp}
Our implementation is based on MatCovNet~\cite{vedaldi15matconvnet}, and publicly available\footnote{\url{https://github.com/bochencaltech/scotopic}}.

In step one of learning, the scalar functions $\alpha(t)$ and $\beta_j(t)$ in~\refeq{eq:approxh} are learned as follows. As the inputs to the network are preprocessed, the preprocessing steps alter the algebraic form for $\alpha$ and $\beta$. For flexibility we do not impose parametric forms on $\alpha$ and $\beta$, but represent them with piecewise cubic Hermite interpolating polynomials with four end points at PPP$=[.22,2.2,22,220]$ (interpolants coded in log-scale). We learned the adapted weights at these end-points by using a different batch normalization module for each PPP. At test time the parameters of the modules are interpolated to accommodate other PPP levels. 

In step two of learning, we compute $S^H(\bN_t)$ for $50$ uniformly spaced PPPs in log scale, and train thresholds $\tau(t)$ for each PPP and for each $\eta$. A regularizer $0.01\sum_t ||\tau(t)-\tau(t+1)||^2$ is imposed on the thresholds $\tau(t)$ to enforce smoothness. In~\refeq{eq:smooth}, the steepness of Sigmoid $\sigma$ is annealed over $500$ iterations of gradient descent, with initial value $0.5$, a decay rate of $0.99$ and a floor value of $0.01$.

\subsection{Results}\label{sec:results}

The speed versus accuracy tradeoff curves in the INT regime are shown in~\reffig{fig-SATwaldnet}{a,b}.
Median PPP versus accuracy tradeoffs for all models in the FR regime are shown in~\reffig{fig-SATwaldnet}{c,d}. All models use constant thresholds for producing the tradeoff curves. In~\reffig{fig-SATwaldnet-finetuned-mean}{a} are average PPP versus accuracy curves when the models use optimized dynamic thresholds described in~\refsec{sec:learning}, step-two. 

\vspace{-.1in}
\subsubsection{Model comparisons}
\vspace{-.1in}

Overall, WaldNet performs well under lowlight. It only requires $< 1$ PPP to stay within $0.1\%$ (absolute) degradation in accuracy on MNIST and around $20$ PPP to stay within $1\%$ degradation on CIFAR10. 

{\it WaldNet is sufficient}. The ensemble was formed using specialists at logarithmically-spaced exposure times, thus its curve is discontinuous in the interrogation regime (esp. \reffig{fig-SATwaldnet}{b}). The peaks delineate transitions between specialists. The ensemble's performance at the specialized light levels $[.22,2.2,22,220]$ also provides a proxy for the performance upper bound by ConvNets of the same architecture (apart from overfitting and convergence issues during learning). Using this proxy we see that even though WaldNet uses $1/4$ the parameters of the ensemble, it stays close to the performance upper bound. Under the FR regime, WaldNet is indistinguishable from the ensemble in MNIST and superior to the ensemble in lowlight conditions ($\leq 22$ PPP, perhaps due to overfitting) of CIFAR10. This showcases WaldNet's ability to handle images at multiple PPPs without requiring explicit parameters. 

{\it Training with scotopic images is necessary}. The photopic classifier retrofitted to lowlight applications performs well at high light conditions ($\geq 220$ PPP) but works poorly overall in both datasets. Investigation reveals that the classifier often stops evidence collection prematurely. This shows that despite effective learning, training with scotopic images and having the proper stopping criterion remain crucial. 


{\it Weight adaptation is necessary}. The rate classifier slightly underperforms WaldNet in both datasets. Since the two system have the same degrees of freedom and differ only in how the first layer feature is computed, the comparison highlights the advantage of adopting time-varying features (\refeq{eq:approxh}).

\vspace{-.2in}
\subsubsection{Effect of threshold learning}
The comparison above under the FR regime uses constant thresholds on the learned log posterior ratios (\reffig{fig-SATwaldnet}{c,d}). Using learned dynamic thresholds (step two of~\refsec{sec:learning}) we see consistent improvement on the {\it average} PPP required for given error rate across all models (\reffig{fig-SATwaldnet-finetuned-mean}{b}), with more benefit for the photopic classifier. \reffig{fig-SATwaldnet-finetuned-mean}{c} examines the PPP histograms on CIFAR10 with constant (FR) versus dynamic threshold (optimized FR). We see with constant thresholds many decisions are made at the PPP cutoff of $220$, so the median and the mean are vastly different. Learning dynamic thresholds reduces the variance of the PPP but make the median longer. This is ok because the Bayes risk objective (\refeq{eq:bayesrisk}) concerns the average PPP, not the median. Clearly which threshold to use depending on whether the median or the mean is more important to the application. 

\subsubsection{Effect of interrogation versus free-response}\label{sec:fr}
\vspace{-.1in}
Cross referencing \reffig{fig-SATwaldnet}{a,b} and~\reffig{fig-SATwaldnet}{c,d} reveals that FR with constant thresholds often brings $3$x reduction in median photon counts. Dynamic thresholds also produce faster {\it average} and {\it median} responses.

\vspace{-.2in}
\subsection{Sensitivity to sensor noise}\label{sec:noise}
How robust is the speedup observed in~\refsec{sec:results} affected by sensor noise? For MNIST and CIFAR10, we take WaldNet and vary independently the dark current, the read noise and the fixed pattern noise. \bc{We also introduce a rotational jitter noise to investigate the model's robustness to motion. The jitter applies a random rotation to the camera (or equivalently to the object being imaged by a stationary camera) where the rotation at PPP follows a normal distribution: $\Delta \theta \sim \mathcal{N}(0, \left(\frac{\sigma_\theta PPP}{220}\right)^2)$, where $\sigma_\theta$ controls the level of jitter. e.g. $\sigma_\theta=22^\circ$ means that at $PPP=220$, the total amount of rotation applied to the image has an std of $22^\circ$. The result is shown in~\reffig{fig-SATwaldnet-noise}{a,b}.}

First, the effect of dark current and fixed pattern noise is minimal. Even an $11\%$ dark current (i.e. photon emission rate of the darkest pixel is $10\%$ of that of the brightest pixel) merely doubles the exposure time with little loss in accuracy. The multiplicative fixed pattern noise does not affect performance because WaldNet in general makes use of very few photons.
Second, current industry standard of read noise ($\sigma_r = 22\%$~\cite{fossum2011quanta}) guarantees no performance loss for MNIST and minor loss for CIFAR10, suggesting the need for improvement in both the algorithm and the photon-counting sensors. The fact that $\sigma_r = 50\%$ hurts performance also suggests that single-photon resolution is vital for scotopic vision. \bc{Lastly, while WaldNet provides certain tolerance to rotational jitter, drastic movement ($22^\circ$ at $220$ PPP) could cause significant drop in performance, suggesting that future scotopic recognition systems and photon-counting sensors should not ignore camera / object motion. }

\subsection{Efficiency of spiking implementation}\label{sec:exp-spiking}
Finally, we inspect the power efficiency of the spiking network implementation (\refeq{eq:rnn},{\bf~\ref{eq:snn}}) on the MNIST dataset. Our baseline implementation (``Continuous'') runs a ConvNet from end-to-end every time the input is refreshed. As a proxy for power efficiency we use the number of multiplications~\cite{merolla2014million}, normalized by the total number in the baseline. For simplicity we vary the discretization threshold $\tau_{dis}$ for inducing spiking events (\refeq{eq:snn}), and the threshold is common across all layers. 

\begin{figure}[t]
\begin{centering}
 \includegraphics[width=1\linewidth]{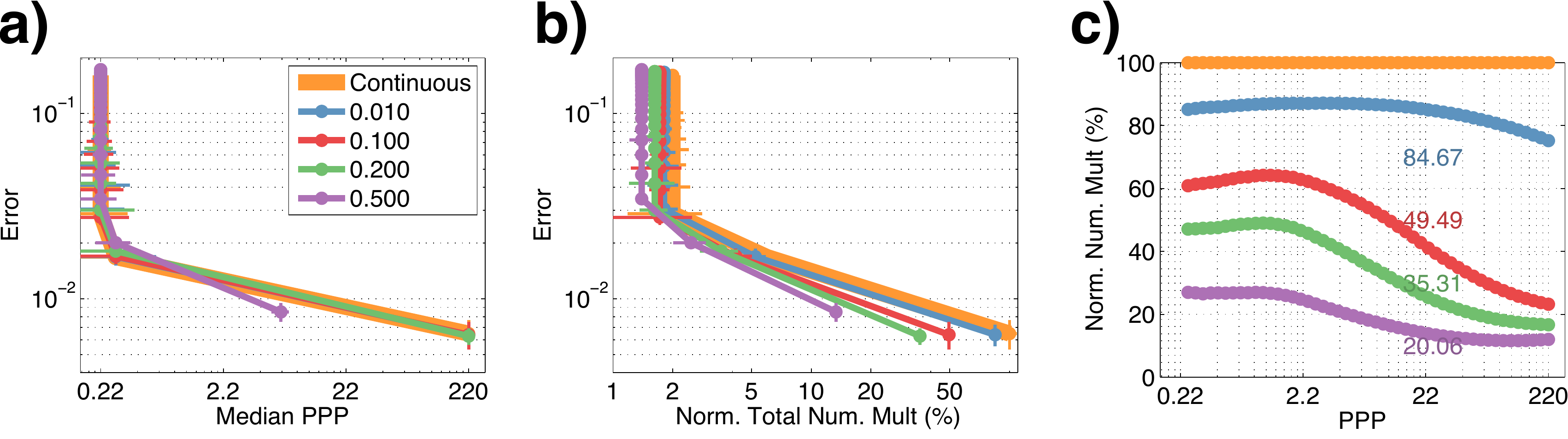}
 \vspace{-0.2in}
\caption{ {\bf ``Power'', speed and accuracy tradeoff of the spiking recurrent neural network implementation on MNIST}. {\bf a)} SAT of spiking networks with different discretization thresholds $\tau_{dis}$ (\refeq{eq:rnn},\refeq{eq:snn}). ``Continuous'' denotes the non-spiking reference implementation. {\bf b)} Error rate as a function of the total amount of multiplications in the network. {\bf c)} The amount of multiplications as a function of time / PPP. Numbers inset represent the average over PPP. }
\label{fig-snn}
\end{centering}
\vspace{-.2in}
\end{figure}

The power, speed and accuracy results shown in~\reffig{fig-snn}{a,b} suggest that for $\tau_{dis} \leq 0.2$ discretization not only faithfully preserves the SAT of WaldNet (\reffig{fig-snn}{a}), but also could be optimized to consume only $35\%$ of the total multiplications, i.e. the spiking implementation provides a $3\times$ power reducation. The amount of spiking events starts high and tappers off gradually (\reffig{fig-snn}{c}) as the noise in the hidden unit estimates (\refeq{eq:rnn}) improves over time. Thus most of the computational savings comes at the later stage ($PPP\geq 22$). Further savings may reside in optimizing the discretization thresholds per layer or over time, which we reserve for future investigations. 

\section{Discussion and Conclusions}
\label{sec-conclusions}
\vspace{-0.1in}

We proposed to study the important yet relatively unexplored problem of scotopic visual recognition. Scotopic vision is vision starved for photons. This happens when available light is low, and image capture time is longer than computation time. In this regime vision computations should start as soon as the shutter is opened, and algorithms should be designed to process photons as soon as they hit the photoreceptors. While visual recognition from limited evidence has been studied~\cite{crouzet2010fast}, 
to our knowledge, our study is the first to explore the exposure time versus accuracy trade-off of visual classification, which is essential in scotopic vision.

We proposed WaldNet, a model that combines photon arrival events over time to form a coherent probabilistic interpretation, and make a decision as soon as sufficient evidence has been collected. The proposed algorithm may be implemented by a deep feed-forward network similar to a convolutional network. Despite the similarity of architectures, we see clear advantages of approaches developed specifically for the scotopic environment. An experimental comparison between WaldNet and models of the conventional kind, such as photopic approaches retrofitted to lowlight images and ensemble-based approaches agnostic of lowlight image statistics, shows large performance differences, both in terms of model parsimony and response time (measured by the amount of photons required for decision at desired accuracy). \bc{WaldNet further allows for a flexible tradeoff between power / computational efficiency with accuracy when implemented as a recurrent spiking network. When trained assuming a constant illuminance, WaldNet may be applied in environments with varying and unknown illuminance levels.} Finally, despite relying only on few photons for decisions, WaldNet is minimally affected by camera noises, making it an ideal model to be integrated with the evolving lowlight sensors. 


{\footnotesize
\bibliographystyle{splncs}
\bibliography{scotopicvision}
}

\appendix
\numberwithin{equation}{section}

\section{Appendix}

\subsection{Time-Adaptation of Hidden Features (\refeq{eq:approxh})} \label{S-app:approxh}
Here we derive the approximation of the first layer activations $S^H(\bN_t)$ given photon count image up to time $t\Delta$ (\refeq{eq:approxh}), copied as below:

\begin{align}
	S^H(\bN_t) &\approx \alpha(t)\boldsymbol{W}\bN_t + \boldsymbol{\beta}(t) 
\end{align}

Recall that we put a Gamma prior on the photon emission rate $\lambda_i$ at pixel $i$:
\begin{align}
	P(\lambda_i) &= Gam(\mu_i \tau, \tau)
\end{align}
where $\mu_i$ is the prior mean rate at pixel $i$. 

After observing $N_{t,i}$ of pixel $i$ in time $[0,t\Delta]$, the posterior estimate for the photon emission rate is:
\begin{align}
	P(\lambda_i | N_{t,i})&\propto P(N_{t,i}|\lambda_i)P(\lambda_i) \\
	&= Gam(  \mu_i\tau + N_{t,i}, \tau + t)
\end{align}
which has a posterior mean of:
\begin{align}
	\hat{\lambda_i} \defas \mathbb{E}[\lambda_i | N_{t,i}] = \frac{\mu_i\tau + N_{t,i}}{\tau+t} \label{S-eq:mean-rate}
\end{align}
Intuitively, the emission rate is estimated via a smoothed-average of the observed counts. Collectively the expected photon counts $\Delta \bN$ over all pixels and duration $(t\Delta, T\Delta)$ given the observed photons $\bN_t$ are:
\begin{align}
	\mathbb{E}[\Delta \bN|\bN_t] = \frac{\boldsymbol{\mu} \tau + \bN_t}{\tau + t} (T-t)
\end{align}
where $\boldsymbol{\mu}$ is the mean rate vector of all pixels. 

Therefore $S^H(\bN_t)$ may be approximated up to second order accuracy using:
\begin{align}
	S^H(\bN_t) &= \sum_{\Delta \bN} (\boldsymbol{W}(\bN_t + \Delta \bN)+\boldsymbol{b}^H) P(\Delta \bN|\bN_t) \\
	&\approx\boldsymbol{W}(\bN_t + \mathbb{E}[\Delta \bN|\bN_t])+\boldsymbol{b}^H \\
	&=\boldsymbol{W}(\bN_t + \frac{\boldsymbol{\mu} \tau + \bN_t}{\tau + t} (T-t))+\boldsymbol{b}^H \\
	&= \underbrace{ \frac{T+\tau}{t+\tau} }_{\alpha(t)} \boldsymbol{W}\bN_t + \underbrace{ \tau\frac{T-t}{\tau+t}\boldsymbol{W}\boldsymbol{\mu} + \boldsymbol{b}^H }_{\boldsymbol{\beta}(t)} 
\end{align}
which proves~\refeq{eq:approxh}.

The equation above works for weights $\boldsymbol{W}$ that span the entire image. In ConvNet, the weights are instead localized (e.g. occupying only a $5\times 5$ region), and organized into groups (e.g. the first layer in WaldNet for CIFAR10 uses $32$ features groups). For simplicity we assume that the mean image $\boldsymbol{\mu}$ is translational invariant within $5\times 5$ regions, so that we only need to model one scalar $\beta_j(t)$ for each feature map $W_j$. 

\subsection{Learning dynamic threshold for Bayes risk minimiziation (\refeq{eq:R_recursive})}\label{S-app:fine-tuning}
Here we show how thresholds $\tau_\eta(t)$ relate to Bayes risk (\refeq{eq:bayesrisk}) in the free-response regime with a cost of time $\eta$. The key is to compute $R^{(n)}_t$, the cumulative future risk from time $t$ for the $n$-th example $\bN_t^{(n)}$ with label $C^{(n)}$. At every point in time, the classifier first incurs a cost $\eta$ (assuming time unit of $1$) in collecting photons for this time point. Then the classifier either report a result according to $S_c(\bN_t^{(n)})$, incurring a lost when the predicted label is wrong, or decides to postpone the decision till later, incurring lost $R^{(n)}_{t+1}$. Which one of the two paths to take is determined by whether the max log posterior crosses the dynamic threshold $\tau_\eta(t)$. Therefore, let $c^* = \arg\max_c S_c(\bN_t^{(n)})$ be the class with the maximum log posterior, the recursion is:
\begin{align}
	R^{(n)}_t &= \eta\Delta + \mathbb{I}[S_{c^*}(\bN_t^{(n)})>\tau_\eta(t)] \mathbb[c^*\neq C^{(n)}] 
	+ \mathbb{I}[S_{c^*}(\bN_t^{(n)})\leq \tau_\eta(t)] R^{(n)}_{t+1} \\
	&= \eta\Delta + q^{(n)}_t e^{(n)}_t + (1-q^{(n)}_t) R^{(n)}_{t+1}
\end{align}
and we assume that a decision must be taken after finite amount of time, i.e. $R^{(n)}_\infty = \eta\Delta + e^{(n)}_\infty$. This proves~\refeq{eq:R_recursive}.

\subsection{Spiking recurrent neural network implementation} \label{S-app:snn}
Here we show that the recurrent dynamics described in~\refeq{eq:rnn} implements the approximation of the first hidden layer activations in~\refeq{eq:approxh}. The proof is constructive: assume that at time $(t-1)\Delta$, the membrane potential $V(t-1)$ computes $S^H(\bN_{t-1})$, i.e. $V(t-1) =\alpha(t-1)\boldsymbol{W}\bN_{t-1} + \boldsymbol{\beta}(t-1)$, then the membrane potential at time $t\Delta$ satisfies:
\begin{align}
V(t) &= r(t) \boldsymbol{V}(t-1) + \alpha(t)\boldsymbol{W} \bX_t + \boldsymbol{l}(t) \\
&= r(t) \left(\alpha(t-1)\boldsymbol{W}\bN_{t-1} + \boldsymbol{\beta}(t-1)\right) + \alpha(t)\boldsymbol{W} \bX_t + \boldsymbol{\beta}(t)-r(t)\boldsymbol{\beta}(t-1) \\
&= \frac{\alpha(t)}{\alpha(t-1)} \alpha(t-1)\boldsymbol{W}\bN_{t-1}+ \alpha(t)\boldsymbol{W}\bX_t + \boldsymbol{\beta}(t) \\
&=\alpha(t)\boldsymbol{W}(\bN_{t-1}+\bX_t) + \boldsymbol{\beta}(t) =\alpha(t)\boldsymbol{W}\bN_t + \boldsymbol{\beta}(t) = S^H(\bN_t)
\end{align}
Hence proving~\refeq{eq:rnn}. 

\subsection{Relationship between exposure time and number of bits of signal ({\bf Table \ref{table-bits}})}
\label{S-app:time-bits}
Bits of signal and photon counts are equivalent concepts. Furthermore, that photon counts are linearly related to exposure time. Here to derive the relationship between exposure time and the number of bits of signal. To simplify the analysis we will make the assumption that our imaging setup has a constant aperture.

What does it mean for an image to have a given number of bits of signal? Each pixel is a random variable reproducing the brightness of a piece of the scene up to some noise. There are two main sources of noise: the electronics and the quantum nature of light. We will assume that for bright pixels the main source of noise is light. This is because, as will be clear from our experiments, a fairly small number of bits per pixel are needed for visual classification, and current image sensors and AD converters are more accurate than that. 

According to the Poisson noise model (\refeq{eq:noise} in main text), each pixel receives photons at rate $\lambda$. The expected number of photons collected during a time $t$ is $\lambda t$ and the standard deviation is $\sigma=\sqrt{\lambda t}$. We will ignore the issue of quantum efficiency (QE), i.e. the conversion rate from photons to electrons on the pixel's capacitor, and assume that QE=1 to simplify the notation (real QEs may range from 0.5 to 0.8). Thus, the SNR of a pixel is $SNR = \lambda t / \sqrt{\lambda t} = \sqrt{\lambda t}$ and the number of bits of signal is $b = \log_2 \sqrt{\lambda t} = 0.5 \log_2 \lambda + 0.5 \log_2  t$.

The value of $\lambda$ depends on the amount of light that is present. This may change dramatically: from $10^{-3}$ LUX in a moonless night to $10^5$ LUX in bright direct sunlight. With a typical camera one may obtain a good quality image in a well lit indoor scene (${\rm E}_v\approx$ 300 lux)  with an exposure time of 1/30s. If a bright pixel has 6.5 bits of signal, the noise is $2^{-6.5} \approx 1\%$ of the dynamic range and $\lambda t / \sqrt{\lambda t} = 100$, i.e. $\lambda \approx  3 \cdot 10^5 \approx 10^3 {\rm E}_v \approx 2^{10} {\rm E}_v $.  Substituting this calculation of $\lambda$ into the expression derived in the previous paragraph we obtain $b \approx 5 + \frac{1}{2} \log_2 t + \frac{1}{2} \log_2 {\rm E}_v$, which is what we used to generate {\bf table~\ref{table-bits}} in the main text.

\end{document}